\documentclass[oribibl]{llncs}
\usepackage{multicol}
\usepackage{amsmath}
\usepackage{amssymb}
\usepackage{graphicx}
\usepackage{epsfig}
\usepackage{color}
\usepackage{hhline}
\usepackage{hyperref}
\usepackage{float}
\usepackage{floatflt}

\newcommand{\ignore}[1]{}

\ignore{+++++++++++++++++++++++++++++
\newcounter{theorem-counter}
\newcounter{corollary-counter}
\newcounter{lemma-counter}
\newcounter{definition-counter}
\newcounter{example-counter}
\newcounter{proposition-counter}
\newcounter{remark-counter}

\setcounter{theorem-counter}{0}
\setcounter{corollary-counter}{0}
\setcounter{lemma-counter}{0}
\setcounter{definition-counter}{0}
\setcounter{example-counter}{0}
\setcounter{proposition-counter}{0}
\setcounter{remark-counter}{0}

{\vskip \abovedisplayskip \refstepcounter{theorem-counter}%
\noindent {\bf Theorem \arabic{theorem-counter}.}}%
{\boxtheorem}

{\vskip \abovedisplayskip \refstepcounter{corollary-counter}%
\noindent {\bf Corollary \arabic{corollary-counter}.}}%
{\boxtheorem}

{\vskip \abovedisplayskip \refstepcounter{lemma-counter}%
\noindent {\bf Lemma \arabic{lemma-counter}.}}%

\newenvironment{definition}%
{\vskip \abovedisplayskip \refstepcounter{definition-counter}%
\noindent {\bf Definition \arabic{definition-counter}.}}%
{\newline}

\newenvironment{example}%
{\vskip \abovedisplayskip \refstepcounter{example-counter}%
\noindent {\bf Example \arabic{example-counter}.}}%

{\vskip \abovedisplayskip \refstepcounter{proposition-counter}%
\noindent {\bf Proposition \arabic{proposition-counter}.}}%

{\vskip \abovedisplayskip \refstepcounter{remark-counter}%
\noindent {\bf Remark \arabic{remark-counter}.}}%

\newcounter{lemmaA-counter}
\newcounter{propositionA-counter}
\setcounter{lemmaA-counter}{0}

\setcounter{propositionA-counter}{0}

{\vskip \abovedisplayskip \refstepcounter{lemmaA-counter}%
\noindent {\bf Lemma A.\arabic{lemmaA-counter}.}}%

{\vskip \abovedisplayskip \refstepcounter{propositionA-counter}%
\noindent {\bf Proposition A.\arabic{propositionA-counter}.}}%

+++++++++++++++++++++}

\newcommand{\boxtheorem}{\hfill $\Box$\\}
\newcommand{\nit}[1]{{\it #1}}
\newcommand{\ul}[1]{\underline{#1}}

\newcommand{\mc}[1]{\mathcal{ #1}}

\newcommand{\mbf}[1]{\mathbf{ #1}}

\newcommand{\mf}[1]{\mathfrak{ #1}}

\newcommand{\e}{\mathbf{e}}
\newcommand{\st}{\mbox{\bf \sf s}}
\newcommand{\doo}{\mbox{\bf \sf do}}
\newcommand{\bstar}{\mathbf{\star}}
\newcommand{\ori}{\mbox{\bf \sf o}}

\title{
 {\bf  An ASP-Based Approach to Counterfactual Explanations for Classification}}

\author{\bf Leopoldo Bertossi\thanks{Faculty  Eng. \& Sciences, Santiago, Chile.  leopoldo.bertossi@uai.cl. \ Member of RelationalAI's Academic Network, and the Millenium Institute on Foundations of Data (IMFD, Chile).}\\Universidad Adolfo Iba\~nez}

\institute{}

\begin{document}
\maketitle
\thispagestyle{empty}
\pagestyle{plain}

\begin{abstract}We propose answer-set programs that specify and compute counterfactual interventions as a basis for causality-based explanations to decisions produced by classification models. They can be applied with black-box models and models that can be specified as logic programs, such as rule-based classifiers. The main focus is on the specification and computation of maximum responsibility causal explanations. The use of additional semantic knowledge is investigated.
\end{abstract}

\section{Introduction}

Providing explanations to results obtained from machine-learning
models has been recognized as critical in many applications, and
has become an active research direction in the broader area of {\em explainable AI}, and explainable machine learning, in particular \cite{molnar}. This
becomes particularly relevant when decisions are automatically made by those
models, possibly with serious consequences for stake holders. Since
most of those models are algorithms
learned from training data, providing explanations may not
be easy or possible. These models are or can be seen as  {\em black-box
models}. \ignore{Furthermore, even when the details of the model are
accessible, there is nothing like a universally accepted definition of
what is an explanation for an outcome of the algorithm.}

In AI, explanations have been investigated in several  areas, and in particular, under {\em actual causality} \cite{Halpern05}, where {\em counterfactual interventions}
on a causal model are central \cite{pearl}. They are hypothetical updates on the model's
variables, to explore if and how the outcome of the
model changes or not. In this way, explanations for an original output are defined and computed. Counterfactual interventions have been  used with ML models, in particular with classification models  \cite{martens,wachter,russell,karimi,datta,lund,deem}.

In this work we introduce the notion of {\em causal explanation} as a set of feature value for the entity under classification that is {\em most responsible} for the outcome. The responsibility score is adopted and adapted from  the general notion of responsibility  used in actual causality \cite{Chockler04}. Experimental results with the responsibility score, and comparisons with other scores  are reported in \cite{deem}. We also introduce {\em answer-set programs} (ASPs) that specify counterfactual interventions and causal explanations, and allow to specify and compute the responsibility score. The programs can be applied with black-box models, and with rule-based classification models.

As we show in this work, our declarative approach to counterfactual interventions is particularly appropriate for bringing into the game additional declarative semantic knowledge, which is much more complicated to do with purely procedural approaches. In this way, we can combine logic-based specifications, and use the generic and optimized solvers behind ASP implementations.

This paper is structured as follows. Section \ref{sec:causes} introduces the background, and the notions of {\em counterfactual intervention} and {\em causal explanation}; and the  {\em explanatory responsibility  score}, $\mbox{\sf x-resp}$, on their basis. Section \ref{sec:casps} introduces ASPs that specify causal explanations, the {\em counterfactual ASPs}. Section \ref{sec:sem} argues for  the need to include semantic domain knowledge in the specification of causal explanations. Section \ref{sec:disc} discusses several issues raised by this work and possible extensions.

\section{Counterfactual Explanations}\label{sec:causes}

We consider {\em classification models}, $\mc{C}$, that are  represented by an input/output relation. Inputs are the so-called {\em entities}, $\e$, which  are represented each by a record (or vector), $\e = \langle x_1, \ldots, x_n\rangle$, where $x_i$ is the value $F_i(\e) \in \nit{Dom}(F_i)$ taken  on $\e$ by a {\em feature} $F_i \in \mc{F} = \{F_1, \ldots, F_n\}$, a set of functions. The output is represented by a {\em label function} $L$ that maps entities $\e$ to $0$ or $1$, the binary result of the classification. That is, to  simplify the presentation, we concentrate here on binary classifiers, but this is not essential.  \ We  also concentrate on features whose domains $\nit{Dom}(F_i)$ take a finite number of categorical values. C.f. Section \ref{sec:sem} for the transformation of numerical  domains into categorical ones.

Building a classifier, $\mc{C}$, from a set of training data, i.e. a set  of pairs $T = \{\langle \e_1,c(\e_1)\rangle,$ $ \ldots, \langle \e_M,c(\e_M)\rangle\}$, with $c(\e_i) \in \{0,1\}$, is one of the most common tasks in machine learning \cite{flach}. It is about learning the label function $L$ for the entire domain of values, beyond $T$. We say that $L$ ``represents" the classifier $\mc{C}$.

Classifiers may take many different internal forms. They could be decision trees, random forests, rule-based classifiers, logistic regression models, neural network-based  (or deep) classifiers, etc. \cite{flach}. Some of them are more ``opaque" than others, i.e. with a more complex and less interpretable internal structure and results \cite{rudin}. Hence the need for explanations to their classification outcomes. In this work, we are not assuming that we have  an explicit classification model, and we do not need it. All we need is to be able to invoke and use it. It could be a ``black-box" model.

The problem is the following: Given an entity $\mbf{e}$ that has received the label $L(\mbf{e})$, provide an ``explanation" for this outcome. In order to simplify the presentation, and without loss of generality, {\em we  assume that label $1$ is the one that has to be explained}. It is the ``negative" outcome one has to justify, such as the rejection of a loan application.

Causal explanations are defined in terms of counterfactual interventions that simultaneously change
feature values in $\e$ in such a way that the updated record gets a new label. A {\em causal explanation} for the classification of
$\e$ is then a set of its original feature values that are affected by a {\em minimal counterfactual interventions}.  These explanations are assumed to be more informative than others. Minimality can be defined in different ways, and we adopt an abstract approach, assuming  a partial order relation $\preceq$ on counterfactual interventions.

\begin{definition} \em \label{def:causal:explanation}  Consider a binary classifier represented by its label function $L$, and a fixed input record $\e = \langle x_1, \ldots, x_n\rangle$, with $F_i(\e) = x_i$, \ $1 \leq i \leq n$, and $L(\e) =1$.

\noindent (a) \ An {\em intervention}  $\iota$ on $\e$ is a set of the form $\{\langle F_{i_1},x_{i_1}'\rangle, \ldots, \langle F_{i_K},x_{i_K}'\rangle\}$, with $F_{i_s} \neq F_{i_\ell}$, for $s\neq \ell$,  \ $x_{i_s} \neq x_{i_s}' \in \nit{Dom}(X_{i_s})$. \ We denote with $\iota(\e)$ the record obtained by applying to $\e$ intervention $\iota$, i.e. by replacing in $\e$ every $x_{i_s} = F_{i_s}(\e)$, with $F_{i_s}$ appearing in $\iota$, by $x_{i_s}'$.

\noindent (b) A {\em counterfactual intervention} on $\e$ is an intervention $\iota$ on $\e$ such that $L(\iota(\e)) = 0$. \ A {\em $\preceq$-minimal} counterfactual intervention is such that there is no counterfactual intervention $\iota'$ on $\e$ with
$\iota' \prec \iota$ \ (i.e. \ $\iota' \preceq \iota$, but not  $\iota \preceq \iota'$).

\noindent (c) A {\em causal explanation} for $L(\e)$ is a set of the form $\epsilon = \{\langle F_{i_1},
x_{i_1}\rangle, \ldots,$  $ \langle F_{i_K}, x_{i_K}\rangle\}$ for which there is a counterfactual intervention $\iota = $ $\{\langle F_{i_1},
x_{i_1}'\rangle, \ldots,$ $ \langle F_{i_K},x_{i_K}'\rangle\}$ for $\e$. \ Sometimes,  to emphasize the intervention, we denote the explanation with $\epsilon(\iota)$.

\noindent (d) A causal explanation $\epsilon$ for $L(\e)$ is {\em $\preceq$-minimal} if it is of the form $\epsilon(\iota)$ for  a $\preceq$-minimal counterfactual intervention $\iota$ on $\e$. \boxtheorem
\end{definition}

\vspace{-5mm}Several minimality criteria can be expressed in terms of partial orders, such as: \ (a) \ $\iota_1 \leq^s \iota_2$ iff $\pi_{1}(\iota_1) \subseteq \pi_{1}(\iota_2)$, with $\pi_{1}(\iota)$ the projection of $\iota$ on the first position. \ (b) \ $\iota_1 \leq^c \iota_2$ iff $|\iota_1| \leq |\iota_2|$.  That is, minimality
under set inclusion and cardinality, resp. \ignore{ A maybe more exotic one when feature values are numerical: \ $\iota_1 \leq^{\mbox{\scriptsize sum}} \iota_2$ iff $\Sigma_{\langle i_j;
x_{i_j},x_{i_j}'\rangle \in \iota_1} |x_{i_j} - x_{i_j}'| \leq  \Sigma_{\langle i_j;
x_{i_j},x_{i_j}'\rangle \in \iota_2} |x_{i_j} - x_{i_j}'|$.} \ In the following, we will consider only these; and mostly the second.

\begin{example} \label{ex:first} Consider three binary features, i.e. $\mc{F} = \{F_1, F_2, F_3\}$, and they take values $0$ or $1$; and the  input/output relation of a classifier $\mc{C}$ shown in Table 1. Let $\e$ be $\e_1$ in the table. We want causal explanations for its label $1$. Any other record in the table can be seen as the result of an intervention on $\e_1$.  However, only $\e_4, \e_7, \e_8$ are (results of) counterfactual interventions in that they switch the label to $0$.

\begin{multicols}{2}
\begin{center}{\Large $\mc{C}$}\\\vspace{1mm}
\begin{tabular}{|c|| c|c|c||c|}\hline
entity (id) & $F_1$ & $F_2$ & $F_3$ & $L$\\ \hline
$\e_1$ & 0 & 1 & 1 &1\\ \hline
$\e_2$ & 1 & 1 & 1 &1\\
$\e_3$ & 1 & 1 & 0 &1\\
$\e_4$ & 1 & 0 & 1 &0\\
$\e_5$ & 1 & 0 & 0 &1\\
$\e_6$ & 0 & 1 & 0 &1\\
$\e_7$ & 0 & 0 & 1 &0\\
$\e_8$ & 0 & 0 & 0 &0\\ \hline
\end{tabular}\\ \vspace{1mm}
{\bf Table 1}
\end{center}

For example, $\e_4$ corresponds to the intervention $\iota_4 = \{\langle F_1,1\rangle, \langle F_2,0\rangle\}$ in that $\e_4$ is obtained from $\e_1$ by changing the values of $F_1, F_2$ into $1$ and $0$, resp. For $\iota_4$,  $\pi_{1}(\iota_4) = \{\langle F_1\rangle,\langle F_2\rangle\}$. From $\iota_4$ we obtain the
 causal explanation $\epsilon_4 = \{\langle F_1, 0\rangle, \langle F_2, 1\rangle\}$, telling  us that the values $F_1(\e_1) = 0$ and $F_2(\e_1) = 1$ are the joint cause for $\e_1$ to have been classified as $1$. There are three causal explanations: \ $\epsilon_4 := \{\langle F_1,0\rangle,\langle F_2,1\rangle\}$, $\epsilon_7 := \{\langle F_2,1\rangle\}$,

\end{multicols}

\vspace{-3mm} \noindent
   and $\epsilon_8 := \{\langle F_2,1\rangle, \langle F_3,1\rangle\}$.   \ Here, $\e_4$ and $\e_8$ are incomparable under $\preceq^s$, \ $\e_7 \prec^s \e_4$, \ $\e_7 \prec^s \e_8$, and  $\epsilon_7$ turns out to be $\preceq^s$- and $\preceq^c$-minimal (actually, minimum). \boxtheorem
\end{example}

\vspace{-3mm}Notice that, by taking a projection, the partial order $\preceq^s$ does not care about the values that replace the original feature values, as long as the latter are changed. \ Furthermore, given $\e$, it would be good enough to indicate the features whose values are relevant, e.g. $\epsilon_7 = \{F_2\}$ in the previous example. However, the introduced notation emphasizes the fact that the original values are those we concentrate on when providing explanations.

Clearly, every $\preceq^c$-minimal explanation is also $\preceq^s$-minimal. However, it is easy to produce an example showing that a $\preceq^s$-minimal explanation may not be $\preceq^c$-minimal. \ignore{
Until further notice we will concentrate only on $\preceq^s$- and $\preceq^c$-minimal explanations; actually more on the latter.}

\ignore{\begin{definition} \em Given a binary classifier represented by its label function $L$, and a fixed input record $\e$: \
(a) An {\em s-explanation} for $L(\e)$ is a $\preceq^s$-minimal causal explanation for $L(\e)$. \
(b) A {\em c-explanations} $L(\e)$ is a $\preceq^s$-minimal causal explanation for $L(\e)$. \boxtheorem
\end{definition} }

\vspace{1mm}\noindent {\bf Notation:} \ An {\em s-explanation} for $L(\e)$ is a $\preceq^s$-minimal causal explanation for $L(\e)$. \
A {\em c-explanations} $L(\e)$ is a $\preceq^c$-minimal causal explanation for $L(\e)$.

\vspace{1mm}
This definition characterizes  explanations as sets of (interventions on) features. However, it is common that one wants to quantify the ``causal strength" of a single feature value in a record representing an entity \cite{lund,deem}, or a single tuple in a database (as a cause for a query answer) \cite{suciu}, or a single attribute value in a database tuple \cite{tocs,foiks18}, etc. \ Different {\em scores} have been proposed in this direction, e.g. {\sf SHAP} in \cite{lund} and {\sf Resp} in \cite{deem}. The latter has it origin in {\em actual causality} \cite{Halpern05}, as the {\em responsibility} of an actual cause \cite{Chockler04}, which we adapt to our setting.

\begin{definition} \em \label{def:er}  Consider  $\e$ to be an entity represented as a record of feature values $x_i=F_i(\e)$, $F_i \in \mc{F}$.

\noindent  (a) \ A feature value $v =F(\e)$, with $F\in \mc{F}$, is a {\em value-explanation}  for $L(\e)$ if there is an s-explanation $\epsilon$ for $L(\e)$, such that $\langle F,v\rangle \in \epsilon$.

\noindent (b) The {\em explanatory responsibility} of a value-explanation $v =F(\e)$ is:

\centerline{$\mbox{\sf x-resp}_{\e,F}(v):= \nit{max}\{\frac{1}{|\epsilon|} : \epsilon \mbox{ is  s-explanation with } \langle F,v \rangle \in \epsilon\}.$}

\noindent (c) If $v =F(\e)$ is not a value-explanation, $\mbox{\sf x-resp}_{\e,F}(v):=0$.
\boxtheorem
\end{definition}

\vspace{-4mm}Notice that (b) can be stated as \ $\mbox{\sf x-resp}_{\e,F}(v):= \frac{1}{|\epsilon^\star|}$, with \
$ \epsilon^\star = \mbox{\sf argmin} \{|\epsilon|~:~\epsilon$ $\mbox{is s-explanation with } \langle F,v\rangle \in \epsilon\}$.

 Adopting the usual terminology in actual causality \cite{Halpern05}, a {\em counterfactual value-explanation} for $\e$'s classification is a value-explanation $v$ with  $\mbox{\sf x-resp}_\e(v)=1$, that is, it suffices, without company of other feature values in $\e$, to justify the classification. \ Similarly, an {\em actual value-explanation} for $\e$'s classification is a value-explanation $v$ with
$\mbox{\sf x-resp}_\e(v) > 0$. That is, $v$ appears in an s-explanation $\epsilon$, say as $\langle F, v\rangle$, but possibly in company of other feature values. In this case, $\epsilon \smallsetminus \{\langle F,v \rangle \}$ is called a {\em contingency set} for $v$ \ \cite{suciu}. \ It turns out that maximum-responsibility value-explanations appear in c-explanations.

\begin{example} (ex. \ref{ex:first} cont.) \  $\epsilon_7$ is the only c-explanation for entity $\e_1$'s classification. Its value $1$ for feature $F_2$ is a value-explanation, and its explanatory responsibility is
$\mbox{\sf x-resp}_{\e_1,F_2}(1):=1$. \boxtheorem
\end{example}

\section{Specifying Causal Explanations in ASP}\label{sec:casps}

 Entities will be represented by a predicate with $n+2$ arguments $E(\cdot;\cdots;\cdot)$. The first one holds a record (or entity) id (which may not be needed when dealing with single entities). The next $n$ arguments hold the feature values.\footnote{For performance-related reasons, it might be more convenient to use $n$ 3-are predicates to represent an entity with an identifier, but the presentation here would be more complicated.}The last argument holds an annotation constant from the set $\{\ori,\doo,\mathbf{\star},\st\}$. Their semantics will be specified below, by the generic program that uses them.

Initially, a record $\e = \langle x_1, \ldots, x_n\rangle$ has not been subject to interventions, and the corresponding entry in predicate $E$ is of the form $E(\e;\bar{x};\ori)$, with $\bar{x}$ an abbreviation for $x_1,\ldots,x_n$, and constant $\ori$ standing for ``original entity".

When the classifier gives label $1$ to $\e$, the idea is to start changing feature values, one at a time. The intervened entity becomes then annotated with constant $\doo$ in the last argument. When the resulting intervened entities are classified, we may not have the classifier specified within the program. For this reason, the program uses a special predicate $\mc{C}[\cdot;\cdot]$, whose first argument takes (a representation of) an entity under classification, and whose second argument returns the binary label. We will assume this predicate can be invoked by an ASP as an external procedure, much in the spirit of HEX-programs \cite{eiter1,eiter2}. \ Since the original instance may have to go through several interventions until reaching one that switches the label to $0$, the intermediate entities get the ``transition" annotation $\bstar$.
\ This is achieved by a generic program.

\paragraph{The Counterfactual Intervention Program:}
\begin{itemize}
\item[P1.] The facts of the program are all the atoms of the form \ $\nit{Dom}_i(c)$, with $c \in \nit{Dom}_i$, plus the initial entity $E(\e;\bar{f};\ori)$, where $\bar{f}$ is the initial vector of feature values.
\item[P2.] The transition entities are obtained as initial, original entities, or as the result of an intervention: \ (here, $\e$ is a variable standing for a record id)
\begin{eqnarray*}
E(\e;\bar{x};\mf{\star}) &\longleftarrow& E(\e;\bar{x};\ori).\\
E(\e;\bar{x};\mf{\star}) &\longleftarrow& E(\e;\bar{x};\doo).
\end{eqnarray*}
\item[P3.] The program rule specifying that, every time the entity at hand (original or obtained after a ``previous" intervention) is classified with label $1$,  a new value has to be picked from a domain, and replaced for the current value.  The new value is chosen via the non-deterministic ``choice operator", a well-established mechanism in ASP \cite{zaniolo}. In this case, the values are chosen from the domains, and are subject to the condition of not being the same as the current value: 
    \end{itemize}
\begin{eqnarray*}
\mbox{\phantom{ooo}}E(\e;x_1',x_2, \ldots,x_n,\mbox{\bf \sf do}) \vee \cdots \vee E(\e;x_1,x_2, \ldots, x_n',\mbox{\bf \sf do})  \ \longleftarrow \ E(\e;\bar{x};\mf{\star}), \mc{C}[\bar{x};1],\\ \hspace*{4cm} \nit{Dom}_1(x_1'), \ldots, \nit{Dom}_n(x_n'),x_1'\neq x_1, \ldots, x_n'\neq x_n,\\ \hspace*{4cm}  \nit{choice}(\bar{x};x_1'), \ldots, \nit{choice}(\bar{x};x_n').
\end{eqnarray*}
For each fixed $\bar{x}$, $\nit{choice}(\bar{x};y)$ chooses a unique value $y$ subject to the other conditions in the same rule body. The use of the choice operator can be eliminated by replacing each $\nit{choice}(\bar{x};x_i')$ atom by the atom $\nit{Chosen}_i(\bar{x},x_i')$, and defining each predicate $\nit{Chosen}_i$ by  means of ``classical" rules \cite{zaniolo}, as follows:
\begin{eqnarray*}
\nit{Chosen}_i(\bar{x},y) &\leftarrow& E(\e;\bar{x};\mf{\star}),  \mc{C}[\bar{x};1], \nit{Dom}_i(y), y \neq x_i, \nit{not} \ \nit{DiffChoice}(\bar{x},y).\\
\nit{DiffChoice}(\bar{x}, y)  &\leftarrow& \nit{Chosen}_i(\bar{x}, y'), y' \neq y.
\end{eqnarray*}

\begin{itemize}
\item[P4.] The following rule specifies that we can ``stop", hence annotation $\st$, when we reach an entity that gets label $0$:\vspace{-2mm}
$$E(\e;\bar{x};\mbox{\bf \sf s}) \ \longleftarrow \ E(\e;\bar{x};\doo), \mc{C}[\bar{x};0].$$

\item[P5.] We add a {\em program constraint} specifying that we prohibit going back to the original entity via local interventions:

\vspace{2mm}\hspace*{3cm}$\longleftarrow \ E(\e;\bar{x};\doo),E(\e;\bar{x};\ori).$

\vspace{2mm}
\item[P6.] The causal explanations can be collected by means of predicates $\nit{Expl}_i(\cdot;\cdot)$ specified by means of:

\vspace{2mm}\hspace*{1cm} $\nit{Expl}_i(\e;x_i) \ \longleftarrow \ E(\e;x_1,\ldots,x_n;\ori), E(\e;x_1',\ldots,x_n';\st), x_i \neq x_i'$.

\vspace{1mm} Actually, each of these is a value-explanation.
\boxtheorem
\end{itemize}

\vspace{-0.4cm}The program will have several stable models due to the disjunctive rule and the choice operator.  Each model will hold intervened versions of the original entity, and hopefully versions for which the label is switched, i.e. those with annotation $\st$. If the classifier never switches the label, despite the fact that local interventions are not restricted  (and this would be quite an unusual classifier), we will not find a model with a version of the initial entity annotated with $\st$. \ Due to the program constraint in P5., none of the models will have the original entity annotated with $\doo$, because those models would be discarded \cite{leone}.

 Notice that the use of the choice operator hides occurrences of non-stratified negation \cite{zaniolo}. In relation to  the use of disjunction in a rule head, the semantics of ASP, which involves model minimality, makes only one of the atoms in the disjunction true (unless forced otherwise by the program itself).

\begin{example} (ex. \ref{ex:first} cont.) Most of the {\em Counterfactual Intervention Program} above is generic. In this particular example, the have the following facts: \ $\nit{Dom}_1(0),$  $\nit{Dom}_1(1),$ $\nit{Dom}_2(0),  \nit{Dom}_2(1), \nit{Dom}_3(0),  \nit{Dom}_3(1)$ and $E(\e_1;0,1,1;\ori)$, with $\e_1$ a constant, the record id of the first row in Table 1.

In this very particular situation, the classifier is explicitly given by Table 1. Then, predicate $\mc{C}[\cdot;\cdot]$ can be  specified with a set of additional facts:
$\mc{C}[0,1,1;1]$, $\mc{C}[1,1,1; 1]$,
$\mc{C}[1, 1, 0;1]$
$\mc{C}[1, 0, 1;0]$
$\mc{C}[1, 0, 0;1]$
$\mc{C}[0, 1, 0;1]$
$\mc{C}[0, 0, 1;0]$
$\mc{C}[0, 0, 0;0]$.

The stable models of the program will contain all the facts above. One of them, say $\mc{M}_1$, will contain (among others) the facts: \ $E(\e_1;0,1,1;\ori)$ and $E(\e_1;0,1,1;\bstar)$. \ The presence of the last atom activates rule P3., because \linebreak $\mc{C}[0,1,1;1]$ is true (for $\e_1$ in Table 1). New facts are produced for $\mc{M}_1$  (the new value due to an intervention is underlined): $E(\e_1;\ul{1},1,1;\doo),$ $ E(\e_1;\ul{1},1,1;\bstar)$. \ Due to the last fact and the true $\mc{C}[1, 1, 1;1]$,
rule P3. is activated again. Choosing the value $0$ for the second disjunct, atoms
$ E(\e_1;\ul{1},\ul{0},1;\doo),$ $ E(\e_1;\ul{1},\ul{0},1;\bstar)$ are generated.  For the latter, $\mc{C}[1, 0, 1;0]$ is true (coming from $\e_4$ in Table 1), switching the label to $0$. Rule P3 is no longer activated, and we can apply rule P4., obtaining \ $E(\e_1;\ul{1},\ul{0},1;\st)$.

 From rules P6., we obtain as explanations:  $\nit{Expl}_1(\e_1;0), \nit{Expl}_2(\e_1;1)$, showing the values in $\e_i$ that were changed. All this in model $\mc{M}_1$. \ There are other models, and one of them contains $E(\e_1;0,\ul{0},1;\st)$, the minimally intervened version of $\e_1$, i.e. $\e_7$.
\boxtheorem
\end{example}

\vspace{-9mm}\subsection{C-explanations and maximum responsibility}
There is no guarantee that the intervened entities $E(\e;c_1,\ldots, c_n;\st)$ will correspond to c-explanations, which are the main focus of this work. In order to obtain them (and only them), we
add {\em weak program constraints} (WCs) to the program. They  can be violated by a stable model of the program (as opposed to (strong) program constraints that have to be satisfied). However, they have to be violated in a minimal way. \ We use WCs, whose {\em number} of violations have to be minimized, in this case, for $1 \leq i \leq n$:
\begin{equation*}
:\sim \ E(\e;x_1,\ldots,x_n,\ori), E(\e;x_1',\ldots,x_n',\st), x_i \neq x_i'.\footnote{This notation follows the standard in \cite{stan}.}
\end{equation*}
Only the stable models representing an intervened version of $\e$ with a minimum number of value discrepancies with $\e$ will be kept.

In each of these ``minimum-cardinality" stable models $\mc{M}$, we can collect the corresponding  c-explanation for $\e$'s classification as the set \ $\epsilon^{\mc M} = \{\langle F_i, c_i\rangle~|$ $\nit{Expl}_i(\e;c_i) \in \mc{M}\}$. This can be done within a ASP system such as {\em DLV}, which allows set construction and aggregation, in particular, counting \cite{dlv2,leone}. Actually, counting comes handy to obtain the cardinality of  $\epsilon^{\mc M}$. The responsibility of a value-explanation $\nit{Expl}_i(\e;c_i)$ will then be: \
$\mbox{\sf x-resp}_{\e,F_i}(c_i) = \frac{1}{|\epsilon^{\mc M}|}$.

\section{Semantic Knowledge}\label{sec:sem}

Counterfactual interventions in the presence of semantic conditions requires consideration. As the following example shows, not every intervention, or combination of them, may be admissible \cite{jdiq}. It is in this kind of situations that declarative approaches to counterfactual interventions, like the one presented here, become particularly useful.

 \begin{example}  A moving company makes automated hiring decisions based on feature values  in applicants' records of the form  $R=\langle \nit{appCode}, \mbox{\nit{ability to lift}}, \mbox{\it gender},$ $ \mbox{\it weight},\mbox{\it height}, \nit{age}\rangle$. \
 Mary, represented by  $R^\star = \langle 101, 1, F, \mbox{\it 160 pounds}, \mbox{\it 6 feet}, 28\rangle$ applies, but  is denied the job, i.e. the classifier returns: \ $L(R^\star) = 1$. \
  To explain the decision, we can hypothetically change Mary's gender, from $\nit{F}$ into $\nit{M}$, obtaining record $R^{\star\prime}$, for which we now observe $L(R^{\star \prime}) = 0$.  Thus, her value $F$ for {\em gender}  can be seen as a counterfactual explanation for the initial decision.

  As an alternative, we might keep the value of \nit{gender}, and  counterfactually change other feature values. However, we might be constrained or guided by an ontology containing, e.g. the denial semantic constraint \  $\neg (R[2] =1 \wedge R[6]  > 80)$  ($2$ and $6$ indicating positions in the record)  that prohibits someone over 80 to be qualified as fit to lift. \ We could also have a rule, such as \ $ (R[3] = M \wedge R[4] > 100 \wedge R[6] < 70)  \rightarrow R[2] =1$, specifying that men who weigh over 100 pounds and are younger than 70 are automatically qualified to lift weight.

In situations like this, we could add to the ASP we had before: (a) program constraints that prohibit certain models, e.g.  \ $\longleftarrow \ R(\e;x, \mbox{\sf 1}, y, z,u, w;\bstar),$ $w>  {\sf 80}$; \ (b) additional rules, e.g. $R(\e;x, \mbox{\sf 1}, y , z,u, w;\bstar) \longleftarrow R(\e;x, y, \mbox{\sf M}, z,u, w;\bstar),$ $ z > {\sf 100}, w < {\sf 70}$,
 that may automatically generate additional interventions. In a similar way, one could accommodate certain preferences using weak program constraints.  \boxtheorem
\end{example} 

\vspace{-5mm}Another situation where not all interventions are admissible occurs when
features take continuous values, and their domains have to be discretized. The common way of doing this, namely the combination of {\em bucketization and one-hot-encoding}, leads to the natural and necessary imposition of additional constraints on interventions, as we will show. \ Through
bucketization, a feature range is discretized by splitting it into finitely many, say $N$, usually non-overlapping intervals. This makes the feature basically categorical (each interval becoming a categorical value). Next, through one-hot-encoding, the original feature is represented as a vector of length $N$ of indicator functions, one for each categorical value (intervals here) \cite{deem}. In this way, the original feature gives rise to $N$ binary features. For example, if we have a continuous feature ``External Risk Estimate" ($\mbox{\sf ERE}$), its buckets could be: $[0, 64), [64, 71), [71, 76), [76, 81),$ $ [81, \infty)$. Accordingly, if
for an entity $\e$,  $\mbox{\sf ERE}(\e) = 65$, then, after one-hot-encoding, this value is represented as the vector $[0, 1, 0, 0, 0, 0]$, because $65$ falls into the second bucket.

In a case like this, it is clear that counterfactual interventions are constrained by the assumptions behind  bucketization and one-hot-encoding. For example, the vector cannot be updated into, say $[0, 1, 0, 1, 0, 0]$, meaning that the feature value for the entity falls both in intervals $[64, 71)$ and $[76, 81)$.
\ Bucketization and one-hot-encoding can make good use of program constraints, such as \ $\longleftarrow \mbox{\sf ERE}(\e; x,1,y,1,z,w;\bstar)$, etc. \ Of course, admissible interventions on predicate $\mbox{\sf ERE}$ could be easily handled with a disjunctive rule like that in P3., but without the ``transition" annotation $\bstar$. However, the $\mbox{\sf ERE}$ record is commonly a component of a larger record containing all the feature values for an entity \cite{deem}. Hence the need for a more general and uniform form of specification.

\section{Discussion}\label{sec:disc}

This work is about interacting with possibly external classifiers and reasoning with their results and potential inputs. That is, the classifier is supposed to have been learned by means of some other methodology. In particular, this is not about learning ASPs, which goes in a different direction \cite{russo}.

We have treated classifiers as black-boxes that are represented by external predicates in the ASP. However, in some cases it could be the case that the classifier is given by a set of rules, which, if compatible with ASPs, could be appended to the program, to define the classification predicate $\mc{C}$. \ The domains used by the programs can be given explicitly. However, they can be specified and extracted from other sources. For example, for the experiments in \cite{deem}, the domains were built from the training data, a process that can be specified and implemented in ASP.

The ASPs we have used are inspired by {\em repair programs} that specify and compute the repairs of a database that fails to satisfy the intended integrity constraints \cite{monica}. Actually, the connection between database repairs and actual query answer causality was established and exploited in \cite{tocs}. ASPs that compute attribute-level causes for query answering were introduced in \cite{foiks18}. They are much simpler that those presented here, because, in that scenario, changing attribute values by nulls is good enough to invalidate the query answer (the ``equivalent" in that scenario to switching the classification label here). Once a null is introduced, there is no need to take it into account anymore, and a single ``step" of interventions is good enough.

Here we have considered only s- and c-explanations, specially the latter. Both embody specific and different, but related, minimization conditions. However, counterfactual explanations can be cast in terms of different optimization criteria \cite{karimi,russell}. One could investigate in this setting other forms on preferences, the generic $\preceq$ in Definition \ref{def:causal:explanation}, by using ASPs as those introduced in \cite{schaub}.
These programs could also de useful to compute (a subclass of) s-explanations, when c-explanations are, for some reason, not useful or interesting enough. The ASPs, as introduced in this work, are meant to compute c-explanations, but extending them is natural and useful.

This article reports on preliminary work that is part of longer term and ongoing research. In particular, we are addressing the following: \ (a)  multi-task classification. \  (b) inclusion of rule-based classifiers. (c) scores associated to more than one intervention at a time \cite{deem}, in particular, to  full causal explanations. \ (d) experiments with this approach and comparisons with other forms of explanations. \ However, the most important direction to explore, and that is a matter of ongoing work, is described next. 

\subsection{From ideal to more practical explanations}

The approach to specification of causal explanations we described so far in this paper is in some sense {\em ideal}, in that the whole product space of the feature domains is considered, together with the applicability of the classifier over that space. This may be impractical or unrealistic. However, we see our proposal as a conceptual and specification basis that can be adapted in order to include more specific practices and mechanisms, hopefully keeping a clear declarative semantics.  One way to go consists in restricting the product space; and this can be done  in different manners. For instance, one can use constrains or additional conditions in rule bodies. An extreme case of this approach consists in replacing the product space with the entities in a {\em data sample} $S \subseteq \Pi_{i=1}^n \nit{Dom}(F_i)$. We could even assume that this sample already comes with classification labels, i.e.
$S^L = \{\langle \e_1',L(\e_1')\rangle, \ldots, \langle \e_K',L(\e_K')\rangle\}$. Actually, this dataset does not have to be disjoint from the training dataset $T$ mentioned early in Section \ref{sec:causes}. The definition of causal explanation and the counterfactual ASPs could be adapted to these new setting without major difficulties.

An alternative and more sophisticated approach consists in using knowledge about the underlying population of entities, such a probabilistic distribution; and using it to define causal explanations, and explanation scores for them. This is the case of the {\sf Resp} and {\sf SHAP} explanation scores mentioned in Section \ref{sec:causes} \cite{deem,lund}. In these cases, it is natural to explore the applicability of probabilistic extensions of ASP \cite{pASP}. In most cases, the underlying distribution is not known, and has to be estimated from the available data, e.g. a sample as $S^L$ above, and the scores have to be redefined (or estimated) through  by appealing to this sample. This was done in \cite{deem} for both {\sf Resp} and {\sf SHAP}. In these cases, counterfactual ASPs could be used, with extensions for set building and aggregations to compute the empirical scores, hopefully in interaction with a database containing the sample.

\vspace{2mm}
\noindent {\bf Acknowledgements:} \ The thorough and useful comments provided by anonymous reviewers are greatly appreciated.

\bibliographystyle{plain}

\end{document}